\documentclass[letterpaper, 10 pt, conference]{ieeeconf} 
\IEEEoverridecommandlockouts
\usepackage[letterpaper, 
            top=0.75in, 
            bottom=1in, 
            left=0.625in, 
            right=0.625in, 
            columnsep=0.25in]{geometry}
            
\usepackage{cite}                  
\usepackage{amsmath,amssymb,amsfonts}
\usepackage{algorithmic}
\usepackage{graphicx}
\usepackage{textcomp}
\usepackage{xcolor}
\usepackage{float}                 
\usepackage[utf8]{inputenc}        
\usepackage{booktabs}              
\usepackage{makecell}
\usepackage{bm}
\usepackage{microtype}             
\usepackage{isomath}               
\usepackage{tabularx}

\usepackage{hyperref}
\hypersetup{
  pdftitle={A Comparative Study of Bayesian Contextual Bandits for Real-Time Warehouse Sorter Optimization},
  pdfauthor={Tina Dongxu Li, Mouhacine Benosman, Ken Meszaros, and Trevor Dardik}
}

\def\BibTeX{{\rm B\kern-.05em{\sc i\kern-.025em b}\kern-.08em
    T\kern-.1667em\lower.7ex\hbox{E}\kern-.125emX}}

\title{A Comparative Study of Bayesian Contextual Bandits for Real-Time Warehouse Sorter Optimization}

\author{Tina Dongxu Li, Mouhacine Benosman, Ken Meszaros, and Trevor Dardik\\
\authorblockA{Amazon.com, USA\\
\{dxl, mbenos, mesza, tdardik\}@amazon.com}
}

\begin{document}
\maketitle 
\thispagestyle{empty} 
\pagestyle{empty}      

\begin{abstract}
Efficient sorter diversion control of automated material handling systems (MHS) is critical for optimizing operational efficiency in large-scale warehouse environments. In this study, we use an inbound receiving sorter at a high-volume e-commerce warehouse as our primary use case, where the sorter diversion system relies on cost functions with static weight configurations that fail to adapt to highly dynamic system contexts, such as volume mode, congestion level, equipment physical status, and upstream/downstream dependencies. To address this real-time sorter diversion optimization challenge, we conducted a comparative study of three candidate hybrid machine learning frameworks: Linear Regression with Gradient Descent Optimization (LR+GDO), XGBoost with Bayesian Optimization (XGB+BO), and Bayesian Contextual Bandits (BCB). Model training and evaluation were enabled by leveraging a high-fidelity physics-aware emulator to overcome the cold-start problem and allow a safe transition from offline to online learning. We performed comprehensive evaluations including reward model predictive accuracy, contextual sensitivity, action distribution, and projected reward uplift. Our results demonstrate that while tree-based reward models offer slightly better predictive power, the BCB framework achieved overall higher performance with $2.03\%$ reward uplift over the heuristic baseline. Furthermore, BCB exhibits several superior characteristics, such as its decisive time-optimal policy backed by Bang-Bang control theory, continuous online learning capability, strategic balance between exploration and exploitation, and significantly shorter inference latency. These results demonstrate the potential of the BCB framework for real-time control optimization in large-scale warehouse environments, motivating further investigation toward operational deployment.

\par\vspace{0.5em}

\end{abstract}

\section{Introduction}

In an e-commerce warehouse environment, various sorters are deployed to manage the diversion of incoming items towards their respective destinations. Each sorter utilizes a customized cost function to decide the diversion direction for incoming items. The cost function calculates a compound cost score for each possible diversion direction by computing a weighted sum of several cost factors in real time. For each incoming item, the sorter evaluates the cost scores across all possible diversion directions and selects the direction with the lowest score. This cost function is tailored to the specific characteristics of each sorter at each site, taking into consideration of factors such as sorter type, site physical layout, number of diversion options, available sensors and system metrics. Although the specific components of each cost function may vary, the general form can be expressed as: 
\begin{equation}
    Compound\ Cost\ Score = \sum_{i=1}^n w_i \cdot cost\_factor_i
\end{equation}
where $cost\_factor_i$ represents the value of the $i$-th cost factor, $w_i$ represents the weight assigned to that cost factor,  and $n$ represents the total number of cost factors. Cost factors are operational metrics that reflect key system performance dimensions. A few examples are destination area fullness, throughput, divert continuity, and assignment preferences, etc.

The warehouse is a highly complex system with interconnected upstream and downstream processes that evolve over time, requiring the sorter diversion system to adapt and make optimal diversion decisions. However, the current limitation is that the weight configurations in the cost functions have been historically static. Fixed configurations can render sub-optimal diversion decisions that lead to high recirculation, congestion, and reduced operational efficiency. To address this issue, we propose a hybrid machine learning framework that automatically recommends optimal weights given a system context. This framework incorporates offline model initialization and online continuous learning stages.

A key challenge in developing such a framework is the cold-start problem. Historical operational data lacks variation in cost weight values and is insufficient to train a robust optimization model. To mitigate this, we leverage a high-fidelity physics-aware emulator. By programmatically assigning random values to the cost weights within the emulator and recording the corresponding system results, we prepare simulated datasets that capture the complex hidden relationships between system dynamics (context), cost weights (action), and resulting performance metrics (reward).

In this paper, we conduct a comparative study of three candidate algorithm architectures for the sorter optimization solution:
\begin{itemize}
    \item Linear Regression + Gradient Descent Optimization (LR+GDO)
    \item XGBoost + Bayesian Optimization (XGB+BO)
    \item Bayesian Contextual Bandit (BCB)
\end{itemize}

Using an inbound receiving sorter at a high-volume fulfillment environment as our primary use case, we collected 5000 training samples by running the emulator. During the offline initialization stage, we train a reward model on these samples to learn the relationship between system state, control actions, and resulting operational outcomes. We then evaluate the trained policy offline by generating actions for held-out system states and using the learned reward model as a surrogate to estimate the corresponding rewards, comparing against a heuristic baseline recorded in the same dataset. The three model candidates are compared based on offline predictive accuracy, simulation results, model complexity, potential for continuous online learning, and real-time inference latency. Ultimately, the results show that the Bayesian Contextual Bandit (BCB) provides the most robust balance of performance and online learning capability for various sorter use cases.

\section{Related Work}
Many existing research about warehouse sorter optimization problems are addressed using traditional Operations Research (OR) methodologies. These works apply heuristic algorithms to make deterministic assignment or scheduling decisions and often utilize Mixed-Integer Linear Programming (MILP) to minimize travel time or maximize throughput, based on known parameters \cite{ZHOU2025, duque2024warehouse, BOYSEN2018386, BOYSEN2024459}. However, in the highly dynamic warehouse sorter environment, simple heuristic rules rarely hold over time as the system evolves on its own and become increasingly stochastic. More importantly, as part of the highly context and interconnected system, the optimal sorter configuration relies on high-dimensional system context inputs which make traditional OR models struggle with the ``curse of dimensionality" and high inference latency which fails to meet the real-time diversion requirement.

Some other studies employ data-driven strategies by training Supervised Learning (SL) models to predict future volume, congestion or equipment failure, followed with rule-based or OR models like linear programming for recommendation \cite{11064978}. Although these models can have high accuracy on predicting certain parameters as inputs for the solver, they are often implemented as open-loop systems that lack the real-time adaptive feedback loop. Our proposed solution builds upon a closed-loop architecture where optimization and learning are tightly integrated.

Contextual Bandits has been successfully applied to many different domains including education (\cite{lan_baraniuk_2016}), health (\cite{liang_xu_taneja_tambe_janson_2024, tewari_murphy_2017}), tourism (\cite{qassimi_rakrak_2025}), and digital marketing such as ad-placement and news recommendation (\cite{li_chu_langford_schapire_2010}). In the industry settings, Bayesian Contextual Bandit (BCB) is emerging as a sample-efficient single-step Bandit model that offers a transparent mechanism for uncertainty quantification via Thompson Sampling, compared to the "black-box" multi-step Deep Reinforcement Learning which is known to be data hungry and requires hundreds of thousands of samples for training \cite{ghavamzadeh_mannor_pineau_tamar_2015}. 

While existing studies have explored using Contextual Bandit for warehouse optimization, they primarily focus on macro logistics and static discrete action planning problems such as order consolidation, picking optimization, and storage allocation \cite{siciliano_braun_zols_fottner_2023}. By contrast, our study moves beyond simple discrete assignments, and uses the Bayesian Contextual Bandit framework to address the continuous cost-weight optimization problem in the critical but under-explored domain of high-frequency warehouse sorter control.

\section{System Modeling: Context Space, Decision Variables, and Objective Functions}

Before we introduce the algorithm designs, we first mathematically formulate this optimization problem that we are addressing. In this section, we define the system context vector, decision variables, and reward function, taking the inbound receiving sorter as an example. 

\subsection{System Context Representation}
The system context can be represented as a numerical vector including features such as
system throughput, upstream scan count, destination area fullness, recirculation rate, control override rate, etc. Some features are included as aggregated sum or mean
over the previous $\Delta T$ rolling window, and some other features are included as time
series that provide temporal information to enhance the model's predictive capability.

\begin{itemize}
    \item Aggregated features include: throughput, recirculation rate,
    control override rate, routing non-compliance rate, destination fullness.
    These features capture the surrounding system dynamics of the sorter, and are aggregated
    over the previous $\Delta T$ time window. Rolling-window aggregation filters out transient noise and provides a stable representation of the true underlying system condition.

    \item Time series features include: upstream scan volume by input category.
    These features provide insights into upstream system throughput, which is useful for
    predicting future sorter arrival volume. Including predicted future arrival volume in
    the context vector is helpful for action recommendation. Rather than building a separate volume prediction model, we directly include raw upstream throughput signals in the context vector, allowing the model to internalize the temporal relationship between upstream volume and future reward.
\end{itemize}

\subsection{Decision Variables}
The decision variables are the weights assigned to the cost factors in the cost function, which is used for making real-time sorter divert decisions. In order to be able to properly attribute the resulting reward to its corresponding cost weights, the cost weights should remain at certain values for a reasonably long duration (e.g. a few minutes) so that the system can react to it and accumulate some reliable impacts on system metrics. Therefore, during the simulation phase, we choose to perturb the values of cost weights every $\Delta$ time window and during the online learning phase, we also have the model to generate new weights recommendations every $\Delta$ time window.

\subsection{Reward Design}

There are several system performance metrics that we would like to improve and balance with this optimization solution. Therefore, we utilize a composite reward score which is defined as a weighted sum of several operational Key Performance Indicators (KPIs):
\begin{equation}
r = \sum_{i=1}^{n} w_i \cdot k_i, \quad \text{subject to } w_i \geq 0,\ \sum_{i=1}^{n} w_i = 1
\label{eq:reward}
\end{equation}
where $k_i$ represents the $i$-th KPI and $w_i$ is its corresponding weight. 
The reward score should be calculated as an aggregated value over a $ \Delta$ time window, rather than a point value collected from a certain timestamp, so that it can reflect the true operation efficiency level of the system and filter out noises. We have analyzed the definition and business logic of each reward metric included in the composite score, the majority of their impacts should come from the immediate cost weights implemented during the same time window. Therefore, the reward observation time window should align with the same time window of its corresponding cost weights. consequently, this model will focus on optimizing the immediate reward, rather than delayed long term reward.

\section{Algorithm Design: Optimization Framework Candidates and Formulations}
This section discusses the algorithm architecture and design details of three candidate optimization frameworks for the sorter divert optimization problem.

\subsection{Linear Regression + Gradient Descent Optimization (LR+GDO)}
The LR+GDO architecture first trains a reward model that captures the complex relationship between system context ($C$), cost weights ($W$) and reward outcomes ($R$). The reward model is trained using linear regression with Lasso regularization to reduce feature dimension and make it a sample efficient baseline. We have included interaction terms $C_i*W_j$ between all context variables and weight variables into the regression model to allow it to capture how cost weights could affect reward differently given different system context.
\begin{equation}
\begin{aligned}
\phi(C, W) 
  &= \bigl[1,\ C_1,\ \ldots,\ C_{d_c},\ W_1,\ \ldots,\ W_{d_w}, \\
  &\quad C_1 W_1,\ \ldots,\ C_{d_c} W_{d_w}\bigr]^\top
\end{aligned}
\label{eq:linear_reward}
\end{equation}
where $d_\phi = 1 + d_c + d_w + d_c \cdot d_w$. The reward model is then formulated as:
\begin{equation}
\hat{R}(C, W) = \beta^\top \phi(C, W)
\end{equation}
The model parameters $\beta$ are learned by minimizing the Lasso-regularized objective:
\begin{equation}
\min_{\beta} \sum_{i=1}^{n} \left(R_i - \beta^\top \phi(C_i, W_i)\right)^2 + \alpha \|\beta\|_1
\end{equation}
where $n$ is the number of training samples and $\alpha$ is the regularization parameter.

Once the reward model is trained, it is used to guide the optimization search for action recommendation. We utilize the PyTorch framework for efficient gradient descent optimization. We extract the learned coefficients from the regression model and port them into a differentiable PyTorch wrapper which reconstructs the reward model within its computation graph, so that it can use this reward model as its objective function to search for global optimum. The optimal solution needs to meet a simplex constraint: 
\begin{equation}
\max_{W} \hat{R}(C, W) \quad \text{subject to} \quad \sum_{i=1}^{d_w} w_i = 1, \quad w_i \geq 0 \quad \forall i
\end{equation}
Given that the default PyTorch optimization search is performed in an unconstrained latent space $Z \in \mathbb{R}^{d_w}$, we add a Softmax layer to map it to the valid weights $W$, then we use Adam optimizer to iteratively update $Z$ by backpropagating the gradient calculated from predicted rewards.

\subsection{XGBoost + Bayesian Optimization (XGB+BO)}
The XGB+BO framework consists of two key components: a tree-based reward model and a Bayesian optimization solver. The XGBoost reward model, initially trained on simulated data, is able to predict system performance outcomes with current context and cost weights.
\begin{equation}
\hat{R}(C, W) = f_{\text{XGB}}([C, W])
\end{equation}
where $[C, W]$ denotes the concatenation of context and weight vectors. 

This reward model is used to guide the optimum search for a real-time optimization solver that dynamically generate weights recommendations as the system context evolves. We chose Bayesian Optimization as our real-time solver due to its distinct advantages: 1) our tree-based reward model is non-differentiable and BO can build a Gaussian process (GP) surrogate model to allow for global optimum search, 2) BO process is very sample efficient and handle such expensive-to-evaluate objective function well, 3) BO provides uncertainty quantification in its predictions. 
Bayesian Optimization utilized a GP process to model the objective function:
\begin{equation}
f_{\text{XGB}}(W) \sim \mathcal{GP}(\mu(W), k(W, W'))
\end{equation}
where $\mu(W)$ is the mean function and $k(W, W')$ is the covariance kernel. We use Matern 2.5 as the kernel function in this study. 
New observations $\mathcal{D}_{1:t} = \{(W^{(i)}, R^{(i)})\}_{i=1}^{t}$ are used to update the posterior distribution:
\begin{equation}
p(R | W, \mathcal{D}_{1:t}) = \mathcal{N}(\mu_t(W), \sigma_t^2(W))
\end{equation}
The optimal weights are found by maximizing reward using an acquisition function:
\begin{equation}
W^{(t+1)} = \arg\max_{W \in \mathcal{W}} \alpha(W | \mathcal{D}_{1:t})
\end{equation}
We use the Upper Confidence Bound (UCB) as the acquisition function $\alpha(W | \mathcal{D}_{1:t})$ in this study.  

As a proposed extension, once the optimization solver is transitioned to an online setting, real-world data can be collected to form a feedback loop and enable continuous learning and adaptability. A separate offline pipeline would be responsible for retraining the XGBoost reward model periodically (e.g., hourly, daily) by gradually replacing simulated data with live operational data. After each retraining cycle, the updated reward model becomes available to the real-time optimization solver to make smarter decisions. This approach enables the reward model to adapt to data distributional shifts and new operational patterns.

\subsection{Bayesian Contextual Bandits (BCB)}

A core advantage of the BCB framework is it inherently balances between exploration and exploitation by learning a probabilistic belief of the reward function instead of point estimates. 
The reward function is in the form of:
\begin{equation}
R = \beta^\top \phi(C, W) + \epsilon, \quad \epsilon \sim \mathcal{N}(0, \sigma^2)
\end{equation}
where $\phi(C, W)$ is the feature vector including interaction terms as defined in equation\eqref{eq:linear_reward}.
The reward function is trained in the form of a Bayesian Linear Regression (BLR) through maintaining a posterior distribution of coefficients: 
\begin{equation}
p(\beta | \mathcal{D}_{1:t}) = \mathcal{N}(\boldsymbol{\mu}_t, \boldsymbol{\Lambda}_t^{-1})
\end{equation} where $\boldsymbol{\mu}_t \in \mathbb{R}^{d_\phi}$ is the posterior mean and $\boldsymbol{\Lambda}_t \in \mathbb{R}^{d_\phi \times d_\phi}$ is the precision matrix. 
Given a batch of new observations, the model update is calculated as  :
\begin{align}
\boldsymbol{\Lambda}_t &= \boldsymbol{\Lambda}_{t-1} + \boldsymbol{\Phi}^\top \boldsymbol{\Phi} \\
\boldsymbol{\phi}_t &= \boldsymbol{\phi}_{t-1} + \boldsymbol{\Phi}^\top \mathbf{R} \\
\boldsymbol{\mu}_t &= \boldsymbol{\Lambda}_t^{-1} \boldsymbol{\phi}_t
\end{align}
where $\mathbf{R} = [R_1, \ldots, R_N]^\top$ is the reward vector, $\boldsymbol{\Phi} \in \mathbb{R}^{N \times d_\phi}$ is the feature matrix, and $\boldsymbol{\phi}_t \in \mathbb{R}^{d_\phi}$ is the accumulated reward-weighted feature vector.

At inference time, BCB uses Thompson Sampling to sample a coefficient vector $\tilde{\beta} \sim \mathcal{N}(\boldsymbol{\mu}_t, \boldsymbol{\Lambda}_t^{-1})$ from the model's current posterior distribution to be used for reward prediction during optimization search. The optimum is defined as \begin{equation}W^* = \arg\max_{W \in \mathcal{W}} \tilde{\beta}^\top \phi(C, W)\end{equation} The context vector is provided as an input, and the action optimization search is performed in an unconstrained latent space $Z \in \mathbb{R}^{d_w}$, and we apply a Softmax transformation to map it to $W$, then we use the calculated softmax Jacobian to iteratively update $Z$ until the reward converges:
\begin{equation}
Z^{(k+1)} = Z^{(k)} + \eta J_{\text{softmax}}(Z)^\top
\nabla_W \left[\tilde{\beta}^\top \phi(C, W)\right]
\end{equation} where $\eta$ is the learning rate and $J_{\text{softmax}}(Z)_{ij} = w_i(\delta_{ij} - w_j)$ is the Softmax Jacobian. The gradient with respect to $W$ is calculated as $
\nabla_W = \beta_W + C^\top B$, where $\beta_W$ are the action coefficients and $B \in \mathbb{R}^{d_c \times d_w}$ are the interaction coefficients. 

The stochastic sampling allows the model to continue to explore weights in the uncertain area (where $\boldsymbol{\Lambda}_t^{-1}$ has large variance) while naturally performing exploitation as the uncertainty shrinks and posterior distribution narrows. Optionally, we can set weight priors using expert domain knowledge to allow efficient early exploration or even add a safety bounds to the search space to guarantee safe exploration. 

Due to this closed-loop design, BCB is able to update its policy in real-time as it continues to collect online feedback and can rapidly adapt to the evolving system dynamics, such as mechanical wear, floor conditions changes, sensor drift, or new operational patterns. 

\section{Simulation Experiment and Comparative Performance Analysis}
Using the inbound receiving sorter at the chosen large-scale e-commerce warehouse as our primary use case, we compare the three candidate optimization frameworks by evaluating their corresponding reward model's accuracy, action distribution, sample efficiency, and projected reward lift returned from high fidelity simulation. 

\subsection{Emulator data collection}
In order to address the cold-start challenge and generate data samples to initialize the reward model training, we leverage a physics-aware emulator that replicates the operational dynamics of the target system. We collect a set of 5000 data samples in the form of $[\text{context},\ \text{action},\ \text{reward}]$ tuples, where each sample captures a snapshot of system state, the applied cost weights, and the resulting operational outcome. In this study, the context vector has dimensionality $d_c = 14$ and the action vector has dimensionality $d_w = 6$, corresponding to the six cost weights in the sorter's cost function. The resulting feature vector $\phi(C, W)$ used in the BCB reward model has dimensionality $d_\phi = 1 + d_c + d_w + d_c \cdot d_w = 105$, including bias, context, action, and all interaction terms. To ensure broad coverage of the action space during emulation, we randomly sample actions from a Dirichlet distribution that satisfies the simplex constraint ($w_i \geq 0,\ \sum_i w_i = 1$). The emulator is configured to execute a different randomly drawn action over each collection interval and record the corresponding system context and reward metrics.
The time window structure is central to the feature engineering and reward calculation process. We define three temporally aligned windows for each sample: 
\begin{itemize}
    \item Context (system conditions): a rolling lookback window ending at decision time $t(0)$
    \item Action (cost weights): applied over a forward execution window $[t(0),\ t(\Delta)]$
    \item Reward (resulting system performance): observed over the window $[t(0),\ t(\Delta)]$
\end{itemize}

\subsection{Reward Model Efficacy Analysis}
\begin{figure}[H]
    \centering
    \includegraphics[width=0.9\linewidth]{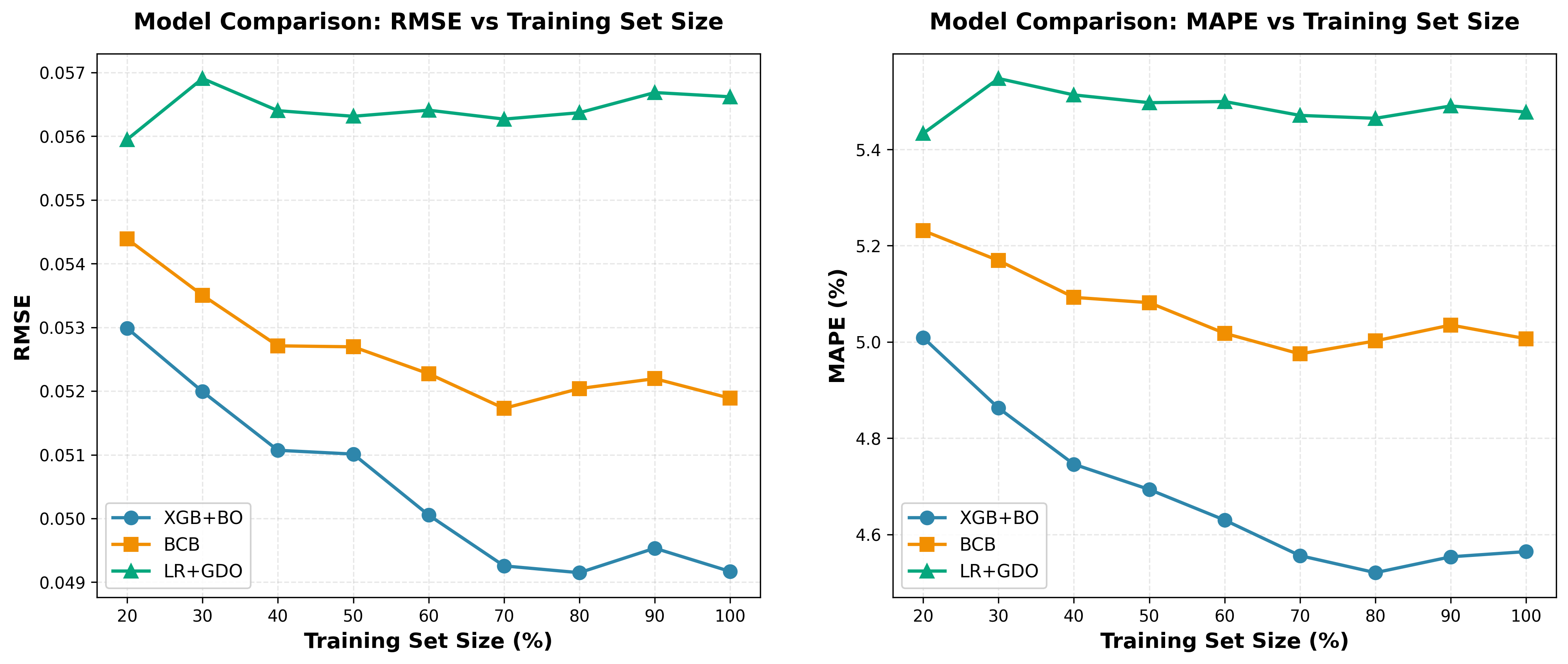}
    \caption{Reward model learning curve comparison}
    \label{fig:learning-curve-comparison}
\end{figure}

All three candidate frameworks have the components of reward prediction and action recommendation. We first evaluate their respective model performance of reward prediction using two metrics: Root Mean Squared Error (RMSE) and Mean Absolute Percentage Error (MAPE). We split data samples as $80\%$ for training and $20\%$ for testing, then we train each model with incremental percentage of training data from $10\%$ to $100\%$ and test them on the same held out data. Finally, we compare the three reward models against a naive mean baseline. As shown in Fig.~\ref{fig:learning-curve-comparison}, both RMSE and MAPE improve as training data size increases for XGB+BO and BCB, with diminishing marginal returns. The curves start to plateau after about 70\% of the data size, suggesting that the models are sample efficient and model performance is converging. By contrast, LR+GDO maintains consistently high error rates across all training sizes, suggesting its inability to effectively learn from data.

Table~\ref{tab:model_summary} presents the final evaluation results on models trained with $100\%$ training dataset. All three ML reward models outperformed the baseline, with the XGB+BO framework achieving the highest performance score, followed by BCB, while LR+GRO yields the poorest results. This performance ranking is consistent across both the RMSE and MAPE metrics. Given LR+GDO's limited learning capacity and consistently inferior performance across both metrics (44.87\% and 48.09\% improvement over baseline, compared to over 49\% for the other two), the remainder of our study focuses exclusively on comparing XGB+BO and BCB, which demonstrate sufficient predictive power to serve as reliable reward surrogates.

\begin{table}[H]
    \centering
    \renewcommand{\arraystretch}{1.2} 
    \caption{Model Prediction Power Summary}
    \label{tab:model_summary}
    \begin{tabular}{l c c c}
        \toprule
        \textbf{Metric} & \textbf{XGB+BO} & \textbf{BCB} & \textbf{LR+GDO} \\
        \midrule
        RMSE             & 0.0492 & 0.0519 & 0.0566\\
        Baseline RMSE    & 0.1027 & 0.1027 & 0.1027 \\
        \textbf{Improvement over baseline}& \textbf{52.12\%} & \textbf{49.50\%} & \textbf{44.87\%}\\
        \midrule
        MAPE             & 4.56\%  & 4.96\%  & 5.48\%\\
        Baseline MAPE    & 10.55\% & 10.55\% & 10.55\% \\
        \textbf{Improvement over baseline}& \textbf{56.74\%} & \textbf{52.98\%} & \textbf{48.09\%}\\
        \bottomrule
    \end{tabular}
\end{table}

\subsection{Feature Importance and Action Sensitivity Analysis}

By analyzing the coefficients of trained BCB reward model and the feature importance of the XGBoost model, we verify two critical assumptions underlying this optimization problem: 1) The cost weights do have impact on reward, 2) The optimal cost weights do vary with system context. The two models are complementary to each other as they describe the physical system from different perspectives, with XGBoost identifying the most influential factors in the system environment, while BCB reveals how the system would respond to cost weights and how the impact of weights would change under different system context .  

Among the top ranking coefficients returned from BCB (Table~\ref{tab:bcb_top_features}), most are interaction terms between context and actions, which confirms that optimal weights are highly sensitive to the system state. For example, the impact of routing-related action weight is conditional on upstream volume indicators and operational performance metrics, with a negative interaction coefficient indicating that the benefit of increasing this weight diminishes during high-load periods. For XGBoost, we utilize SHAP (SHapley Additive exPlanations) to decompose feature impact into main effects and interaction effects, isolating action-context sensitivity. We report a normalized strength score (mean absolute SHAP value relative to the maximum) for each feature in Table~\ref{tab:xgb_top_features}. Both models consistently identify volume and load-related features as the most impactful factors for reward outcomes.  

\begin{table}[H]
\centering
\small
\caption{Top 10 Features from BCB Reward Model}
\label{tab:bcb_top_features}
\begin{tabularx}{\columnwidth}{@{} c >{\raggedright\arraybackslash}X c @{}}
\toprule
\textbf{Rank} & \textbf{Feature} & \textbf{Direction} \\
\midrule
1  & [Upstream throughput context (T-1)] $\times$ [Volume priority weight]       & Negative \\
2  & (Bias)                                                          & Positive \\
3  & [Recirc context] $\times$ [Volume priority weight]       & Positive \\
4  & [Recirc context] $\times$ [Throughput weight]             & Positive \\
5  & [Recirc context] $\times$ [Assignment preference weight]  & Positive \\
6  & [Fullness context] $\times$ [Throughput weight]            & Negative \\
7  & [Fullness context] $\times$ [Volume priority weight]      & Negative \\
8  & Fullness context                               & Positive \\
9  & [Fullness context] $\times$ [Fullness weight]      & Negative \\
10 & [Fullness context] $\times$ [Assignment preference weight] & Negative \\
\bottomrule
\end{tabularx}
\end{table}

\begin{table}[H]
\centering
\small
\caption{Top 10 Features from XGBoost Reward Model (SHAP)}
\label{tab:xgb_top_features}
\begin{tabularx}{\columnwidth}{@{} c >{\raggedright\arraybackslash}X c @{}}
\toprule
\textbf{Rank} & \textbf{Feature} & \textbf{Norm. Strength} \\
\midrule
1  & Context: Routing compliance indicator         & 1.000 \\
2  & Action: Volume priority weight               & 0.819 \\
3  & Context: Destination Fullness                  & 0.497 \\
4  & Context: Upstream throughput (T-1)                     & 0.424 \\
5  & Context: Sorter throughput                 & 0.238 \\
6  & Action: Fullness weight               & 0.207 \\
7  & Context: Control override count          & 0.191 \\
8  & Action: Divert continuity weight         & 0.182 \\
9  & Context: Upstream throughput (T-3)               & 0.157 \\
10 & Context: Destination Fullness        & 0.102 \\
\bottomrule
\end{tabularx}
\end{table}

\subsection{Action Distribution Analysis}
\begin{figure}[H]
    \centering
    \includegraphics[width=0.9\linewidth]{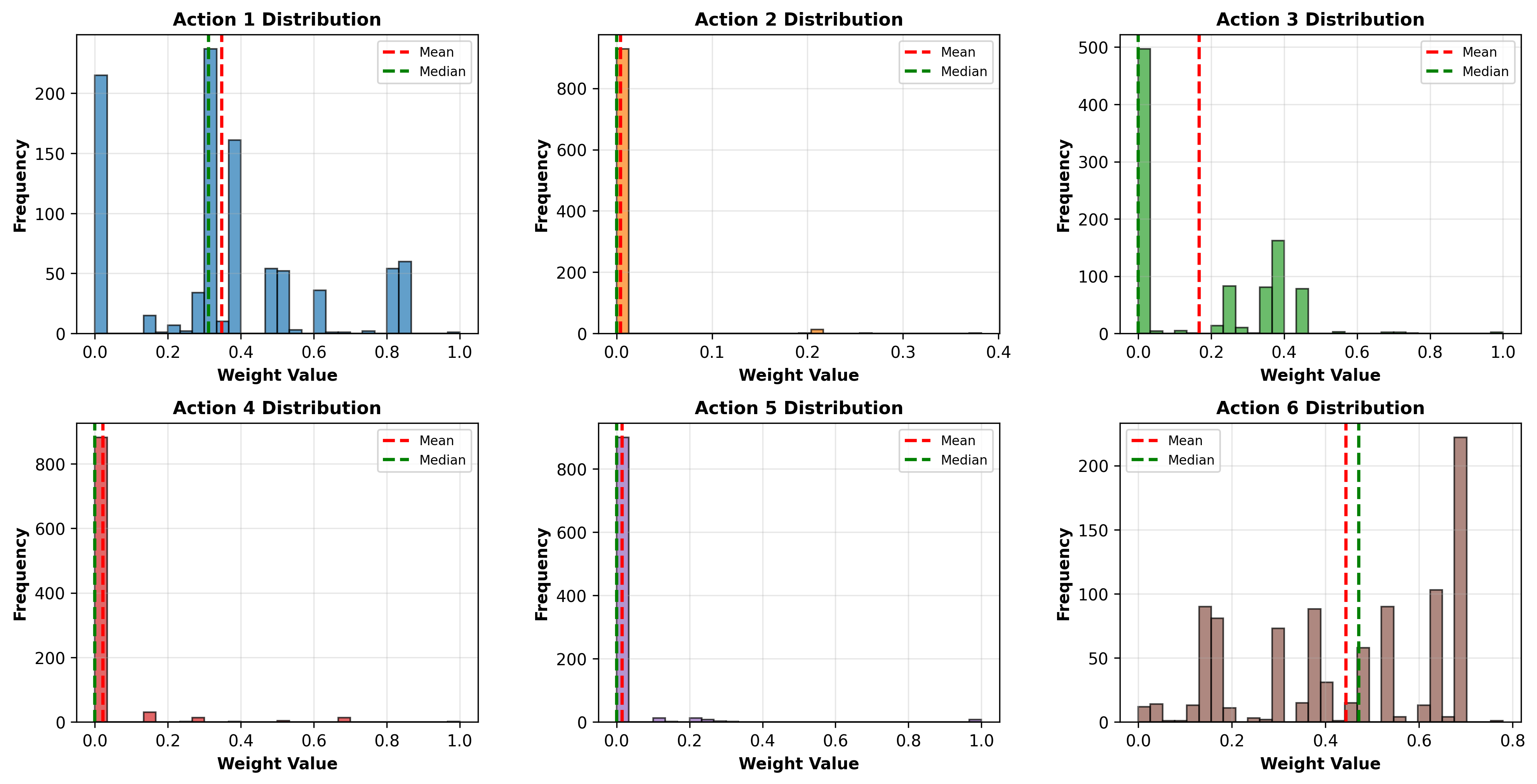}
    \caption{XGB action histogram}
    \label{fig:xgb_action_histogram}
\end{figure}

\begin{figure}[H]
    \centering
    \includegraphics[width=0.9\linewidth]{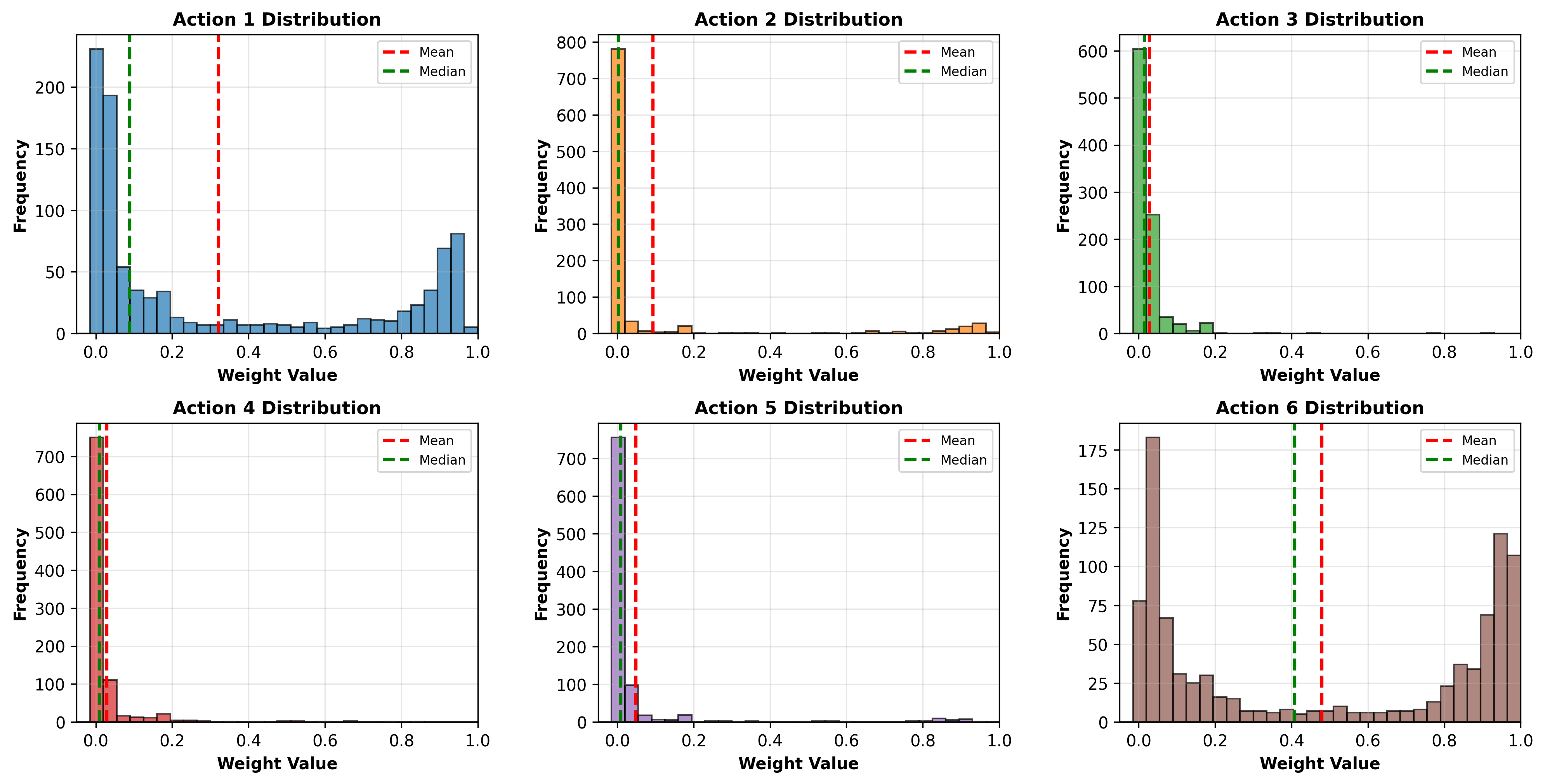}
    \caption{BCB action histogram}
    \label{fig:bcb_action_histogram}
\end{figure}

The histograms in Fig.~\ref{fig:xgb_action_histogram}  and Fig.~\ref{fig:bcb_action_histogram} compare the distributions of actions recommended by XGB+BO versus BCB. Overall, both models effectively learn to assign smaller values to Action 2, 4 and 5. BCB exhibits a “U” shape distribution for Action 1 and 6, identifying specific contexts where the weight must be maxed out to optimize reward, while XGB+BO displays a broader action distribution, favoring the conservative middle-ground. BCB demonstrates to be a more decisive and reactive policy compared to the conservative XGB+BO. 

The bimodal distribution of BCB (switching abruptly between 0 and 1 for high-impact actions like Action 1 and 6) is analogous to Bang-Bang control, a feedback control strategy that switches between extremes rather than intermediate values \cite{sonneborn_vleck_1964}. While a formal optimal-control derivation is beyond the scope of this study, the empirical behavior suggests that decisive corrections may be more effective than moderate adjustments in this high-frequency control setting. Given our frequent model inference cycles, this approach allows BCB to act as a rapid feedback controller that applies maximum effort to correct system deviations as fast as possible before the next inference cycle, instead of applying a gentle adjustment.

\subsection{Reward Uplift Estimation}
To estimate the projected reward uplift, we conducted Python-based simulations to compare candidate frameworks with a heuristic baseline. The heuristic baseline is defined as a fixed weight configuration determined through domain expertise, iterative tuning by operations engineers, and prior operational studies, representing the best-known static policy prior to the introduction of a data-driven optimization framework. This baseline reflects a principled, operationally grounded reference point rather than an arbitrary or naive choice. In each iteration, we provide the models with one context vector from the test dataset, and ask the model to recommend optimal weights, then we estimate corresponding rewards of policy actions and baseline actions using the trained reward model. We calculate performance of policy vs baseline using the identical testing context vectors to ensure a fair comparison. As shown in Table~\ref{tab:reward_uplift}, both models significantly outperform the baseline, but BCB proves to be superior by delivering a 2.03\% increase in reward.    

\begin{table}[H]
    \centering
    \caption{Reward Uplift Comparison}
    \label{tab:reward_uplift}
    \begin{tabular}{cccc}
        \toprule
        \textbf{Model} & \textbf{Reward Uplift (\%)} \\
        \midrule
        XGB+BO  & 1.75\% \\
        BCB     & 2.03\% \\
        \bottomrule
    \end{tabular}
\end{table}

\subsection{Experiment summary and framework recommendation}

Although the XGBoost reward model reported a slightly lower MAPE, the BCB model achieved higher final reward uplift over the heuristic baseline, which is the primary optimization objective. We recommend the BCB framework for real-time sorter diversion optimization, due to its unique advantages summarized as following:
\begin{itemize}
    \item \textbf{Superior online learning capacity}: The warehouse environment is highly dynamic and its underlying system may drift over time as new equipment and software are installed or patterns change. BCB expertly balances exploration and exploitation through Bayesian uncertainty.       
    \item \textbf{Contextual sensitivity}: Our coefficient analysis demonstrates that BCB effectively captures the interactions between context and actions, making optimal weights sensitive to system dynamics. In contrast, XGB's feature importance doesn't prove its policy is able to leverage the nuanced relationship between environment and actions.   
    \item \textbf{Time-optimal control}: BCB’s bimodal action distribution allows for decisive, timely system corrections. 
    \item \textbf{Operational efficiency}: BCB has extremely short inference latency (milliseconds) which facilitates near-instantaneous adaptation. By contrast, the XGB+BO pipeline requires an average of $18$ seconds per inference.
\end{itemize}

\section{Conclusion}
The major contribution of this work is that we mathematically formulated the warehouse sorter environment into a structured Context-Action-Reward representation and conducted a comprehensive evaluation on three end-to-end machine learning frameworks for real-time sorter optimization. The Bayesian Contextual Bandit (BCB) emerged as the superior solution, achieving a 2.03\% reward uplift over the heuristic baseline. Given its empirical alignment with Bang-Bang Control principles, significantly higher computational efficiency, practical sample efficiency and continuous online adaptation capability, the BCB model is the better choice for a high-frequency, context-aware optimization solution required for a highly complex, evolving warehouse environment. Future work will validate the reported reward uplift through closed-loop emulator testing with repeated randomized trials and statistical significance analysis.

\appendix
\section{Appendix}

\subsection{Action Distributions} 
\begin{figure}[H]
    \centering
    \includegraphics[width=0.8\linewidth]{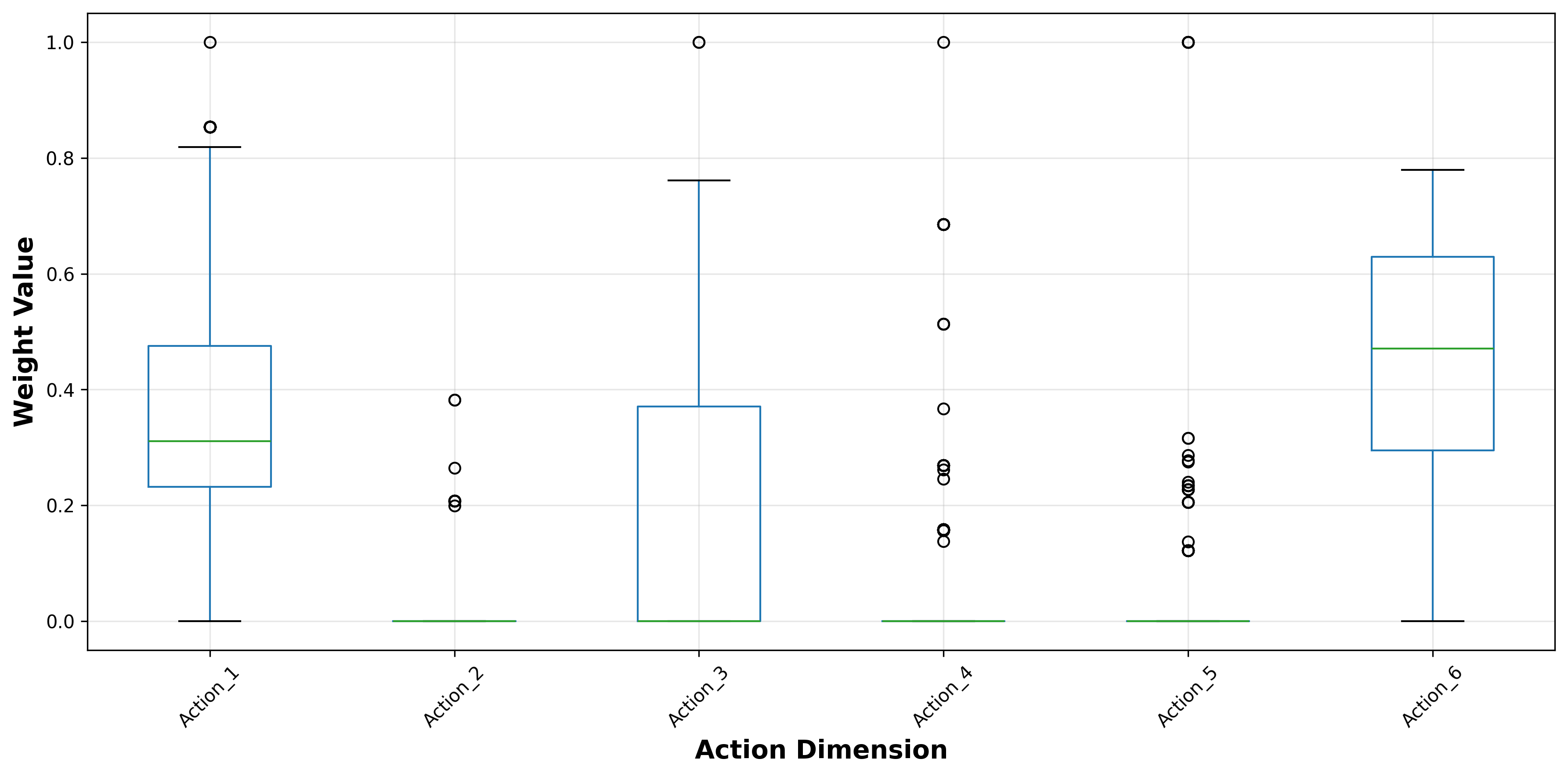}
    \caption{XGB action box plot}
    \label{fig:xgb-action-boxplot}
\end{figure}

\begin{figure}[H]
    \centering
    \includegraphics[width=0.8\linewidth]{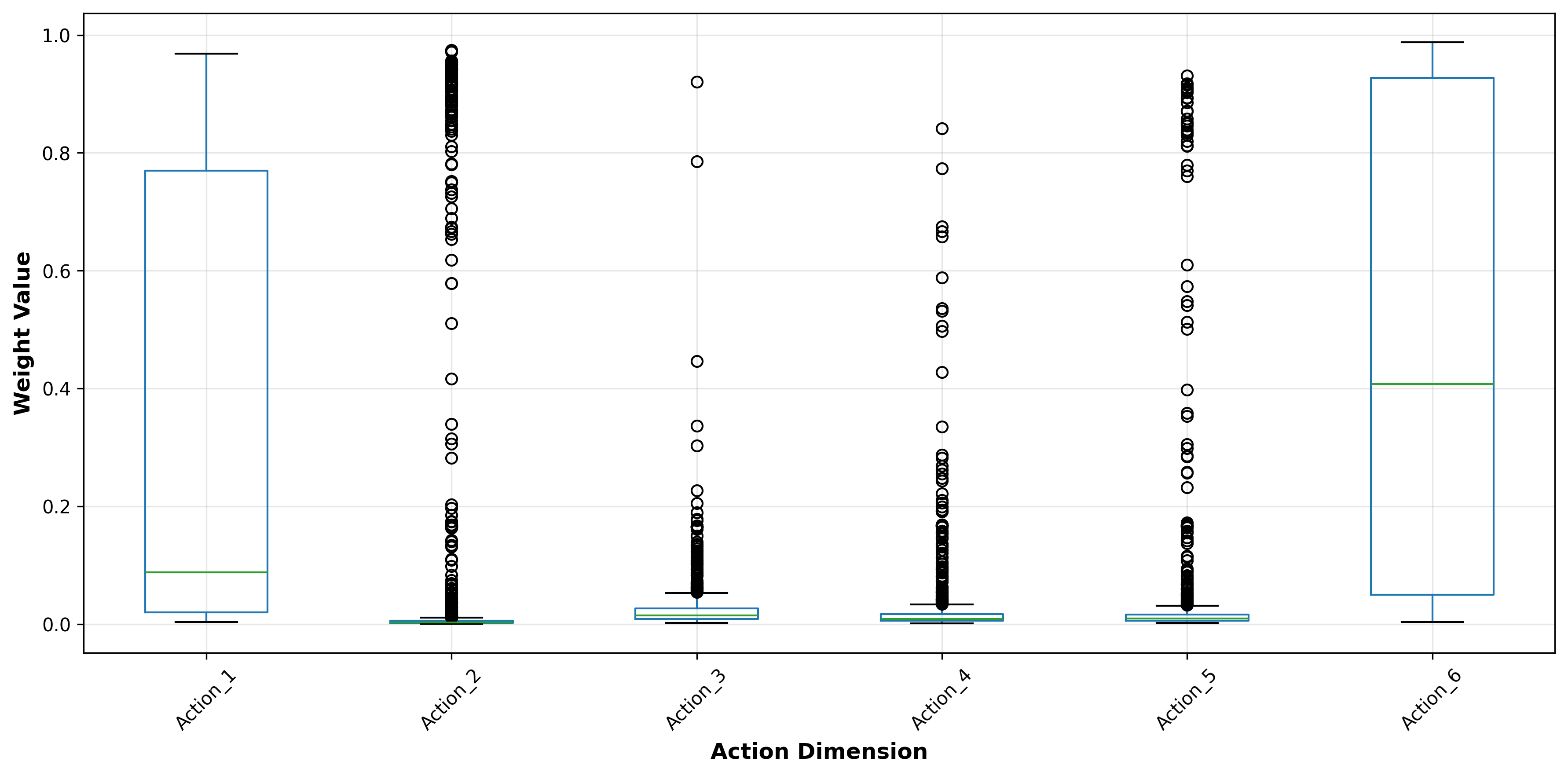}
    \caption{BCB action box plot}
    \label{fig:bcb-action-boxplot}
\end{figure}

\begin{figure}[H]
    \centering
    \includegraphics[width=0.7\linewidth]{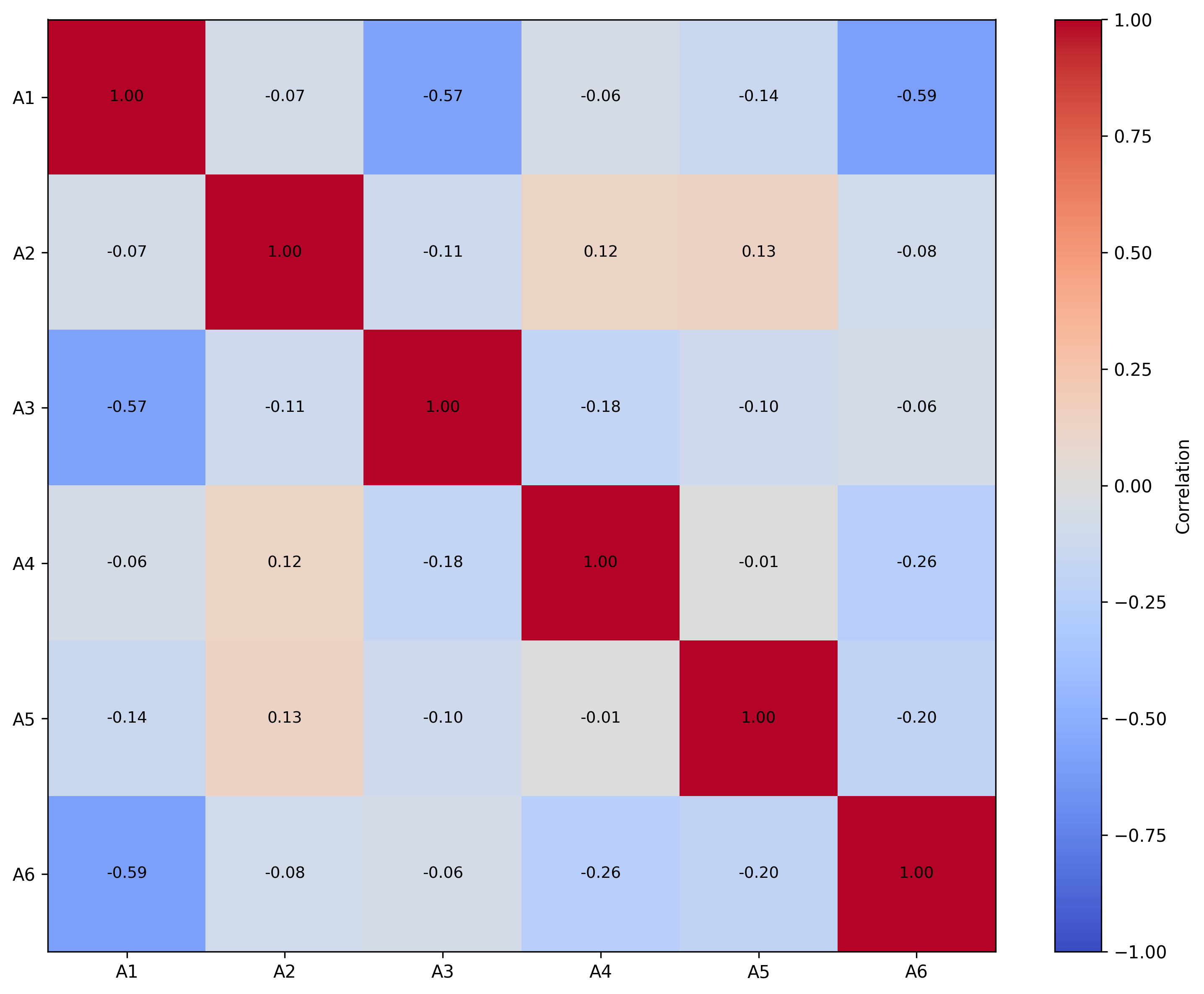}
    \caption{XGB action correlation matrix}
    \label{fig:xgb-action-correlation-matrix}
\end{figure}

\begin{figure}[H]
    \centering
    \includegraphics[width=0.7\linewidth]{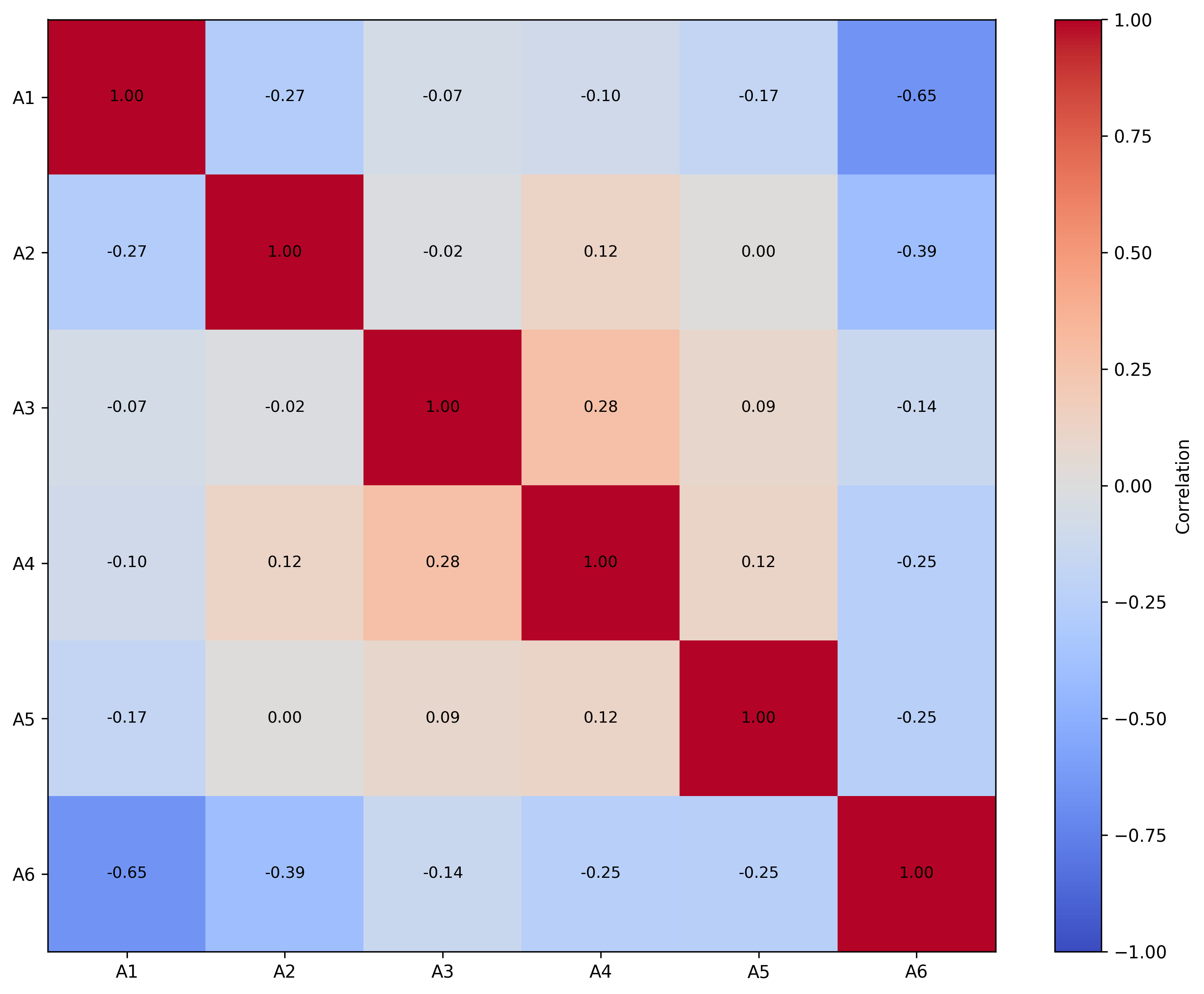}
    \caption{BCB action correlation matrix}
    \label{fig:bcb-action-correlation-matrix}
\end{figure}

Figure~\ref{fig:xgb-action-boxplot}, ~\ref{fig:bcb-action-boxplot}, ~\ref{fig:xgb-action-correlation-matrix}, and ~\ref{fig:bcb-action-correlation-matrix} include the box plots and correlation matrix of recommended actions by XGB+BO versus BCB. 

\section*{ACKNOWLEDGMENT}
The authors would like to thank the leadership team for their support and guidance throughout this project. We also acknowledge the simulation team and software engineering team for their assistance in developing and validating the emulator used for model training and evaluation.

\bibliography{references}
\bibliographystyle{IEEEtran}

\end{document}